\documentclass{article}

\usepackage{PRIMEarxiv}

\usepackage[utf8]{inputenc} 
\usepackage[T1]{fontenc}    
\usepackage{hyperref}       
\usepackage{url}            
\usepackage{booktabs}       
\usepackage{amsfonts}       
\usepackage{nicefrac}       
\usepackage{microtype}      
\usepackage{lipsum}
\usepackage{fancyhdr}       
\usepackage{graphicx}       
\graphicspath{{media/}}     

\usepackage{xspace}
\usepackage{natbib}
\usepackage[dvipsnames]{xcolor}
\usepackage{multirow}
\usepackage{amsmath}
\usepackage{adjustbox}
\usepackage{tcolorbox}
\usepackage{colortbl}
\usepackage{enumerate} 
\usepackage{enumitem}

\newcommand{\datasetshort}{\textsc{Clear-3K}\xspace}
\newcommand{\datasetfull}{\textbf{C}ausal \textbf{L}ogical \textbf{E}xplanatory \textbf{A}ssessment \textbf{R}esource\xspace}
  
\title{\datasetshort: Assessing Causal Explanatory Capabilities \\ in Language Models
}

\author{
  Naiming Liu \\
  Rice University \\
  \texttt{nl35@rice.edu} \\
   \And
  Richard Baraniuk \\
  Rice University \\
  \texttt{richb@rice.edu} \\
  \And
  Shashank Sonkar \\
  University of Central Flordia \\
  \texttt{shashank.sonkar@ucf.edu} \\
}

\begin{document}
\maketitle

\begin{abstract}
We introduce \datasetshort, a dataset of 3,000 assertion-reasoning questions designed to evaluate whether language models can determine if one statement causally explains another. 
Each question present an assertion-reason pair and challenge language models to distinguish between semantic relatedness and genuine causal explanatory relationships. 
Through comprehensive evaluation of 21 state-of-the-art language models (ranging from 0.5B to 72B parameters), we identify two fundamental findings. 
First, language models frequently confuse semantic similarity with causality, relying on lexical and semantic overlap instead of inferring actual causal explanatory relationships. 
Second, as parameter size increases, models tend to shift from being overly skeptical about causal relationships to being excessively permissive in accepting them.  
Despite this shift, performance measured by the Matthews Correlation Coefficient plateaus at just 0.55, even for the best-performing models.
Hence, \datasetshort provides a crucial benchmark for developing and evaluating genuine causal reasoning in language models, which is an essential capability for applications that require accurate assessment of causal relationships.
\end{abstract}


\section{Introduction}
Assertion-reasoning questions have emerged as a valuable tool for evaluating higher-order thinking in educational assessment \citep{kumar2018assessment, cbse2020guidelines}. These questions present students with an assertion followed by a reason, asking them to determine whether the reason correctly explains the assertion. Unlike conventional multiple-choice questions that focuses on factual memorization, assertion-reasoning questions requires the skill of distinguishing between two statements that are merely topically related and those that reflects a genuine explanatory relationship \citep{bloom1956taxonomy, anderson2001revised, sonkar2024malalgoqa}. This distinction represents a fundamental challenge in logical reasoning that requires deeper critical thinking capabilities beyond surface-level topic recognition and memorization \citep{kahneman2011thinking, evans2013dual}.

\begin{figure*}[t!]
    \centering
    \includegraphics[width=\linewidth]{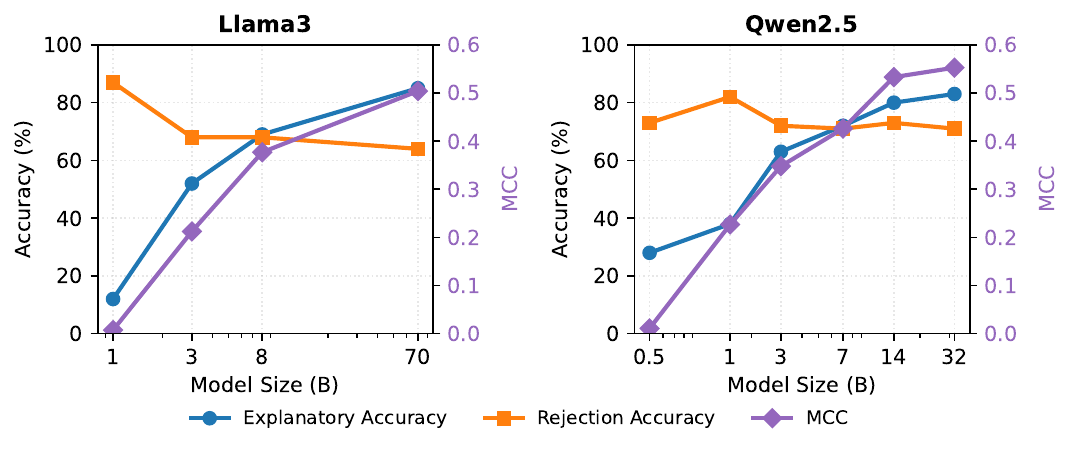}
    \caption{Model performance with increasing sizes. Smaller models exhibit strong negative bias when evaluating explanatory relationship (Llama3-1B: 87\% rejection accuracy, 12\% explanatory accuracy), while larger models develop the opposite tendency (Llama3-70B: 85\% explanatory accuracy, 64\% rejection accuracy). Although overall performance improves with scale, MCC stays at 0.55 across model families. This trend reveals a fundamental limitation: models struggles to distinguish semantic relatedness from genuine causal explanatory relationships.}
    \label{fig:scale-performance}
\end{figure*}

Despite the importance of evaluating causal relationships, current natural language understanding benchmarks inadequately address this reasoning capability. While existing datasets assess various reasoning aspects, including natural language inference \citep{bowman2015large, williams2018broad}, educational comprehension task \citep{rajpurkar2016squad, alag}, and commonsense reasoning \citep{talmor2019commonsenseqa, sakaguchi2020winogrande}, few specifically target the ability to determine whether a given reason constitutes a valid explanation for an assertion. This gap is particularly concerning given the prevalence of causal reasoning in critical domains, from medical diagnosis and treatment decisions to legal argumentation and policy analysis \citep{thorne2018fever, wadden2020fact}.

To address this limitation, we introduce \datasetshort, a dataset of 3,000 assertion-reasoning questions designed to evaluate causal explanatory reasoning capabilities of language models. The questions are curated from real-world educational materials and covers a diverse range of subjects spanning both STEM and humanities disciplines from grades 9 to 12. This diversity allows for robust evaluation across different knowledge domains, difficulty levels, and reasoning styles.

We conduct an extensive empirical evaluation of 21 language models spanning five leading open-source model families, with parameter sizes ranging from 0.5B to 72B parameters. This comprehensive evaluation allows us to identify robust patterns in how causal explanatory reasoning changes with different model types and parameters.

Our primary finding reveals a fundamental limitation in how language models approach causal explanatory reasoning. Language models tend to substitute semantic similarity for causal understanding, regardless of whether a true explanatory relationship exists. This pattern is illustrated in the confusion matrix with semantic similarity statistics from Phi-14B:

\begin{table}[h!]
\centering
\begin{adjustbox}{width=0.5\linewidth}
\begin{tabular}{ccc}
\toprule
& \textbf{Pred Y} & \textbf{Pred N} \\
\midrule
\textbf{True Y} & 0.590 ± 0.073 (1048) & 0.553 ± 0.080 (129) \\
\textbf{True N} & 0.571 ± 0.072 (238) & 0.547 ± 0.083 (480) \\
\bottomrule
\end{tabular}
\end{adjustbox}
\vspace{5mm}
\caption{Confusion matrix for Phi-4-14B when both assertion and reason are true. Y/N indicates whether the reason actually explains the assertion. Reported values show semantic similarity (mean ± std) and counts.}
\label{tab:confusion_matrix}
\end{table}

\begin{table*}[ht!]
    \centering
    \renewcommand{\arraystretch}{1}
    \begin{tabular}{cp{14cm}}
        \toprule
        \rowcolor{blue!10} \multicolumn{2}{p{16cm}}{ \textbf{Given a statement of Assertion (A) followed by a statement of a Reason (R), their relationship falls into one of the four categories:}} \\
        \rowcolor{blue!10} \multicolumn{2}{p{16cm}}{ \hspace{1cm} a. Both A and R are true and R is the correct explanation of A} \\
        \rowcolor{blue!10} \multicolumn{2}{p{16cm}}{ \hspace{1cm} b. Both A and R are true and R is not the correct explanation of A} \\
        \rowcolor{blue!10} \multicolumn{2}{p{16cm}}{ \hspace{1cm} c. A is true but R is false} \\
        \rowcolor{blue!10} \multicolumn{2}{p{16cm}}{ \hspace{1cm} d. A is false but R is true} \\
        \midrule
        \midrule
        \rowcolor{gray!21} \textbf{Category} &  \textbf{Mathematics, Grade 9} \\
        \multirow{2}{*}{\textbf{a}} & \textbf{Assertion:} If angles A and B form a linear pair of angles and A = 70°, then B = 110° \\
        & \textbf{Reason:} Sum of linear pair of angles is always 180° \\
        \midrule
        
        \rowcolor{gray!21} \textbf{Category} &  \textbf{Biology, Grade 10} \\
        \multirow{3}{*}{\textbf{b}} & \textbf{Assertion:} Cerebellum controls the coordination of body movement and posture. \\
        & \textbf{Reason:} Medulla oblongata controls and regulates the centre for coughing, sneezing and vomiting. \\
        \midrule
        
        \rowcolor{gray!21} \textbf{Category} &  \textbf{Physics, Grade 11} \\
        \multirow{3}{*}{\textbf{c}} & \textbf{Assertion:} Total energy of the freely falling body is constant at each point. \\
        & \textbf{Reason:} Kinetic energy of freely falling body is minimum, when it reaches the ground. \\
        \midrule
        
        \rowcolor{gray!21} \textbf{Category} &  \textbf{Chemistry, Grade 12} \\
        \multirow{2}{*}{\textbf{d}} & \textbf{Assertion:} Copper is a non-transition element. \\
        & \textbf{Reason:} Copper has completely filled d-orbitals in its ground state. \\
        \bottomrule
    \end{tabular}
    \caption{Four relationships between assertions and reasons (categories a-d) in assertion-reasoning questions. Each example demonstrates one category, and contains subjects of Mathematics, Biology, Physics, and Chemistry across grades 9-12.}
    \label{tab:dataset_example}
\end{table*}

This results in Table~\ref{tab:confusion_matrix} reveals that when the model correctly identifies causal relationships (True Y, Pred Y), the average semantic similarity is significantly higher ($0.590$) than when it misses the explanations relationships (True Y, Pred N: $0.553$). Similarly, when the model incorrectly predicts explainability for non-explainable pairs (True N, Pred Y), the statements have higher similarity ($0.571$) than when it correctly rejects them (True N, Pred N: $0.547$). These consistent differences suggest that models rely heavily on semantic similarity as a proxy for causality, which is a classic case of confusing correlation with causation. 

Additionally, we observe a consistent shift in model performance as parameter size increases: smaller models exhibit a strong bias toward rejecting explanatory relationships, while larger models develop the opposite trends, as shown in Figure~\ref{fig:scale-performance}. For instance, when both assertion and reason are true, LLaMA3-1B correctly identifies only 12\% of valid explanations while correctly rejecting 87\% of non-explanatory relationships, whereas LLaMA3-70B flips this pattern, achieving 85\% explanatory accuracy and only 64\% rejection accuracy. This pattern suggests that increased model size enhances their sensitivity to semantic relatedness but does not enhance their ability to distinguish genuine causal relationships. To quantify overall performance, we use the Matthews Correlation Coefficient (MCC), which improves with scale across all families but plateaus around 0.55, even for the largest models. This convergence suggests an upper bound on current models’ causal reasoning abilities. 

These results reveal a critical insight about the capabilities of language models: they fundamentally confuse semantic similarity with causal relationships. This limitation persists even as models scale up, suggesting that future progress may require architectural or training innovations specifically targeting causal inference.

\section{Dataset:\datasetshort}
\subsection{Assertion-Reasoning Question Format}
Assertion-reasoning questions are a widely used assessment format in education designed to evaluate students' higher-order thinking abilities. Each question presents two statements - an assertion (A) and a reason (R) — and asks students to classify their relationship into one of the following four categories:

\begin{enumerate}[label=\alph*., itemsep=0pt, parsep=0pt, topsep=0pt, leftmargin=*]
\item Both A and R are true, and Reason correctly explains the Assertion.
\item Both A and R are true, but the Reason does not explain the Assertion.
\item A is true, but R is false.
\item A is false, but R is true.
\end{enumerate}

This format challenges students to not only evaluate the factual accuracy but also determine whether an explanatory causal relationship exists between the assertion and reason statements. Table~\ref{tab:dataset_example} provides representative examples from our dataset illustrating each answer category.

\subsection{Dataset Collection and Composition}

To enable a systematic study of causal reasoning abilities in language models, we present an open-source \datasetshort (\datasetfull), a comprehensive human-curated dataset consisting of 3,008 assertion-reasoning questions spanning multiple domains and difficulty levels. The questions were collected from standardized educational resources, including textbooks, previous examinations, and online educational repositories, with a focus on maintaining diversity across academic subjects and complexity.

Each question in our dataset adheres to the standard assertion-reasoning format described above, comprising an assertion (A), a reason (R), and a classification into one of the predefined four categories (a, b, c, or d). We manually verified the questions to ensure correct answer classifications according to educational standards and subject matter expertise.

\subsection{Subject and Grade Distribution}

Our dataset covers eight distinct subject areas across Grade 9-12, representing both STEM and humanities disciplines as shown in Table \ref{tab:subject_dist}.

\begin{table}[h!]
\centering
\resizebox{0.6\columnwidth}{!}{
\begin{tabular}{lrrrrr}
\toprule
\textbf{Subject} & \textbf{Count} & \textbf{Grade 9} & \textbf{10} & \textbf{11} & \textbf{12} \\
\midrule
Math & 763 & 55 & 263 & 79 & 366 \\
Biology & 605  & 31 & 146 & 104 & 324 \\
Chemistry & 597 & 37 & 150 & 139 & 271 \\
Physics & 576 & 44 & 108 & 126 & 298 \\
Geography & 145 & 28 & 117 & 0 & 0 \\
Pol. Science & 122 & 29 & 93 & 0 & 0 \\
Economics & 100 & 23 & 77 & 0 & 0 \\
History & 100 & 29 & 71 & 0 & 0 \\
\midrule
\textbf{Total} & \textbf{3,008} & \textbf{276} & \textbf{1,025} & \textbf{448} & \textbf{1,259}\\
\bottomrule
\end{tabular}
}
\vspace{5mm}
\caption{Distribution of \datasetshort across subjects and grade levels.}
\label{tab:subject_dist}
\end{table}

The subject distribution reflects the emphasis on STEM disciplines in standard educational curricula. Mathematics, Biology, Chemistry, and Physics collectively comprise of approximately 84\% of the questions in our dataset. In terms of grade level, the dataset is weighted toward Grade 10 and 12, which aligns with the widespread use of assertion-reasoning questions in standardized assessments at these grade levels. Additionally, the humanities subjects (Geography, Political Science, Economics, and History) are only represented in Grade 9 and 10, while STEM subjects span all four grade levels.

\subsection{Answer Category Distribution}

Table \ref{tab:answer_dist} shows the distribution of correct answers for the assertion-reasoning questions across the four categories (a, b, c, d) described above.

\begin{table}[h]
\centering
\resizebox{0.5\columnwidth}{!}{
\begin{tabular}{lcr}
\toprule
\textbf{Category} & \textbf{Count} & \textbf{\%} \\
\midrule
(a) Both true, R explains A & 1,177 & 39.1 \\
(b) Both true, R doesn't explain A & 718 & 23.9 \\
(c) A true, R false & 616 & 20.5 \\
(d) A false, R true & 497 & 16.5 \\
\bottomrule
\end{tabular}
}
\vspace{5mm}
\caption{Distribution of \datasetshort by answer categories.}
\label{tab:answer_dist}
\end{table}

The dataset has a moderate bias toward category (a), which is typical in educational contexts where assertion-reasoning questions are often designed to assess correct understanding of causal relationships. Nevertheless, categories (b), (c), and (d) have sufficient representation to ensure balanced evaluation across different reasoning patterns, with (b) being slightly more common than (c) and (d). This comprehensive dataset with its broad coverage of subjects, grade levels, and answer categories provides a robust foundation for evaluating causal reasoning abilities in language models across diverse knowledge domains.

\section{Problem Formulation}
As described in Section 2.1, assertion-reasoning questions require evaluating both the factual correctness of the assertion and reason statements and determining whether a causal explanatory relationship exists between them. While this format serves educational purposes effectively, it presents challenges when specifically analyzing language models' causal explanatory reasoning capabilities. To address this limitation, we introduce the \textbf{Causal Explanation Task}, a reformulated version that enables more precise evaluation of a model’s capacity to identify genuine explanatory relationships and goes beyond simply recognizing topic relevance or semantic relatedness of two statements.

\subsection{Reformulating Assertion-Reasoning into Causal Explanation Task}

Our primary interest is examining whether language models can correctly identify when one statement (reason) explains another (assertion), which is a fundamental aspect of causal reasoning. Traditional assertion-reasoning questions combine this explanation task with factual verification, which can obscure a model’s causal reasoning and factual verification abilities.

To more precisely evaluate the causal reasoning abilities, we introduce the Causal Explanation Task. This task aims to isolates causal explanatory reasoning as a standalone binary decision problem. This reformulation allows us to assess whether models can go beyond surface-level semantic relatedness and truly identify causal explanatory relationships - a fundamental challenge in logical reasoning for both human and language models. In this binary task format, each question presents a pair of statements, an assertion (A) and a reason (R), and prompts the model to decide whether the reason provides a valid causal explanation for the assertion:

\begin{tcolorbox}[colback=blue!5, colframe=blue!40, sharp corners]
\textbf{Causal Explanation Task}:\\
Assertion (A): \{assertion\}\\
Reason (R): \{reason\}\\
Task:\\
Determine whether the Reason (R) explains the Assertion (A).\\
Provide your answer in the following JSON format:\\
\{"Verdict": "yes" or "no"\}
\end{tcolorbox}

In Causal Explanation task, a "yes" verdict only corresponds to the original option (a), where R truly explains A. For all other cases (b, c, d), the correct verdict is "no" — either because R does not explain A despite both being true (b), or because a factually incorrect statement cannot explain or be explained by another statement (c, d). We evaluate model performance in two distinct settings:

\begin{enumerate}
    \item \textbf{When both A and R are true} (original answers a and b): This context directly tests the model's ability to distinguish between semantic correlation and causation. Even when both statements are factually correct and semantically related, the model must determine whether R actually explains A. This setting directly probes higher-order reasoning that traditional assertion-reasoning questions aim to assess.
    \item \textbf{When either A or R is false} (original answers c and d): This context reveals whether factual errors affect causal reasoning. Logically, a false statement cannot validly explain or be explained by another statement, so models should consistently reject explanatory relationships in this setting.
\end{enumerate}

This task reformulation is crucial because it explores whether models substitute semantic similarity for real causal understanding. By comparing performance across these settings, we can determine whether models simply associate semantically related statements or truly comprehend when one statement causally explains another. As our results will show, models generally perform well in rejecting explanatory relationships when either statement contain factual errors (Setting 2), but struggle to distinguish between semantic correlation and causation when both statements are correct (Setting 1). This pattern suggests that while language models can detect factual inconsistencies, they fundamentally rely on semantic similarity as a proxy for causal relationships, which is a limitation for downstream applications that require rigorous causal reasoning.

\begin{table*}[t]
\centering
\begin{adjustbox}{width=\linewidth}
\begin{tabular}{lccccccc}
\toprule
\multirow{2}{*}{\textbf{Model}} & \multirow{2}{*}{\textbf{Overall MCC}} & \multicolumn{4}{c}{\textbf{When both A, R are true}} & \textbf{When either A or R is false} \\
\cmidrule(lr){3-6} \cmidrule(lr){7-7}
 & & \textbf{Accuracy} & \textbf{MCC} & \textbf{Explanatory} & \textbf{Rejection} & \textbf{Rejection} \\
\midrule
\rowcolor{gray!20} \multicolumn{7}{c}{\textbf{Qwen-3 Models}} \\
\textbf{Qwen3-32B} & \textbf{0.66} & \textbf{0.79} & \textbf{0.55} & 0.85 & 0.69 & 0.90 \\
\textbf{Qwen3-14B} & \textbf{0.66} & 0.78 & \textbf{0.55} & 0.80 & 0.76 & \textbf{0.94} \\
\textbf{Qwen3-8B} & 0.63 & 0.77 & 0.54 & 0.78 & 0.76 & 0.91 \\
\textbf{Qwen3-4B} & 0.62 & 0.77 & 0.52 & 0.81 & 0.71 & 0.88 \\
\textbf{Qwen3-1.7B} & 0.54 & 0.74 & 0.44 & 0.79 & 0.65 & 0.83 \\
\midrule
\rowcolor{gray!20} \multicolumn{7}{c}{\textbf{LLaMA-3 Models}} \\
\textbf{LLaMA3.3-70B} & 0.61 & 0.77 & 0.50 & 0.85 & 0.64 & 0.85 \\
\textbf{LLaMA3.1-8B} & 0.43 & 0.69 & 0.38 & 0.70 & 0.68 & 0.77 \\
\textbf{LLaMA3.2-3B} & 0.30 & 0.59 & 0.21 & 0.53 & 0.68 & 0.80 \\
\textbf{LLaMA3.2-1B} & 0.04 & 0.40 & -0.01 & 0.12 & \textbf{0.87} & 0.93 \\
\midrule
\rowcolor{gray!20} \multicolumn{7}{c}{\textbf{Qwen-2.5 Models}} \\
\textbf{Qwen2.5-72B} & 0.64 & \textbf{0.79} & \textbf{0.55} & 0.87 & 0.66 & 0.87 \\
\textbf{Qwen2.5-32B} & 0.62 & \textbf{0.79} & \textbf{0.55} & 0.84 & 0.71 & 0.85 \\
\textbf{Qwen2.5-14B} & 0.60 & 0.78 & 0.53 & 0.80 & 0.74 & 0.86 \\
\textbf{Qwen2.5-7B} & 0.50 & 0.72 & 0.43 & 0.73 & 0.71 & 0.82 \\
\textbf{Qwen2.5-3B} & 0.42 & 0.67 & 0.35 & 0.64 & 0.72 & 0.81 \\
\textbf{Qwen2.5-1.5B} & 0.23 & 0.55 & 0.23 & 0.39 & 0.83 & 0.82 \\
\textbf{Qwen2.5-0.5B} & 0.02 & 0.45 & 0.01 & 0.27 & 0.73 & 0.74 \\
\midrule
\rowcolor{gray!20} \multicolumn{7}{c}{\textbf{Phi-4 Models}} \\
\textbf{Phi-4-14B} & 0.65 & 0.78 & \textbf{0.55} & 0.78 & 0.78 & 0.92 \\
\textbf{Phi-4-4B} & 0.55 & 0.73 & 0.43 & 0.78 & 0.65 & 0.86 \\
\midrule
\rowcolor{gray!20} \multicolumn{7}{c}{\textbf{Gemma-3 Models}} \\
\textbf{Gemma3-27B} & 0.49 & 0.72 & 0.37 & \textbf{0.93} & 0.36 & 0.67 \\
\textbf{Gemma3-12B} & 0.52 & 0.74 & 0.43 & 0.87 & 0.52 & 0.74 \\
\textbf{Gemma3-4B} & 0.35 & 0.70 & 0.32 & 0.90 & 0.36 & 0.47 \\
\bottomrule
\end{tabular}
\end{adjustbox}
\vspace{2mm}
\caption{
Performance of 21 models from five model families (Qwen-3, LLaMA-3, Qwen-2.5, Phi-4, and Gemma-3) on the CLEAR-3K dataset for the Causal Explanation Task. Results are reported using Explanatory Accuracy, Rejection Accuracy, and MCC across three settings: overall performance, cases where both the A and R are true (assessing the ability to distinguish causal explanation from semantic relatedness), and cases where either A or R is false (assessing whether factual errors affect causal explanatory relationship). Best performance for each metric is highlighted in \textbf{bold}.}
\label{tab:main_results}
\end{table*}

\begin{table*}[t]
\centering
\begin{adjustbox}{width=\linewidth}
\begin{tabular}{c|ccc|ccc}
\toprule
\multirow{2}{*}{\textbf{Model}} & \multicolumn{3}{c|}{\textbf{When both A and R are true}} & \multicolumn{3}{c}{\textbf{When either A or R is false}} \\
\cmidrule{2-7}
 & & \textbf{Pred Y} & \textbf{Pred N} & & \textbf{Pred Y} & \textbf{Pred N} \\
\midrule
\multirow{2}{*}{\textbf{Phi-14B}} & \textbf{True Y} & 0.590 ± 0.073 (1048) & 0.553 ± 0.080 (129) & \textbf{True Y} & -- & -- \\
 & \textbf{True N} & 0.571 ± 0.072 (238) & 0.547 ± 0.083 (480) & \textbf{True N} & 0.567 ± 0.079 (157) & 0.563 ± 0.084 (956) \\
\midrule
\multirow{2}{*}{\textbf{Phi-4B}} & \textbf{True Y} & 0.599 ± 0.069 (851) & 0.552 ± 0.077 (326) & \textbf{True Y} & -- & -- \\
 & \textbf{True N} & 0.583 ± 0.062 (214) & 0.543 ± 0.084 (504) & \textbf{True N} & 0.593 ± 0.066 (200) & 0.557 ± 0.086 (913) \\
\midrule
\multirow{2}{*}{\textbf{Qwen3-14B}} & \textbf{True Y} & 0.593 ± 0.072 (938) & 0.557 ± 0.078 (237) & \textbf{True Y} & -- & -- \\
 & \textbf{True N} & 0.580 ± 0.070 (169) & 0.547 ± 0.082 (546) & \textbf{True N} & 0.566 ± 0.083 (108) & 0.563 ± 0.084 (1004) \\
\midrule
\multirow{2}{*}{\textbf{Qwen3-8B}} & \textbf{True Y} & 0.592 ± 0.073 (919) & 0.564 ± 0.077 (248) & \textbf{True Y} & -- & -- \\
 & \textbf{True N} & 0.581 ± 0.070 (167) & 0.547 ± 0.082 (547) & \textbf{True N} & 0.584 ± 0.073 (121) & 0.560 ± 0.085 (982) \\
\bottomrule
\end{tabular}
\end{adjustbox}
\vspace{2mm}
\caption{Confusion matrices with semantic similarity between A and R (mean ± std with sample counts). Y/N indicates whether the reason correctly explains the assertion. In the setting ``When either A or R is false", True Y cells are empty because false statements cannot logically provide valid explanations.}
\label{tab:confusion_matrices}
\end{table*}

\subsection{Evaluation Metrics}

We evaluate model performance on Causal Explanation task using three complementary metrics. For each assertion-reason pair, language models must determine whether the reason explains assertion by responding with either "yes" (indicating R provides a valid causal explanation for A) or "no" (indicating R does not causally explain A). Based on these binary verdicts, we measure:

1. \textbf{Explanatory Accuracy}: The proportion of cases in which the models correctly respond "yes" when R truly explains A (corresponding to original answer (a)). This metric measures the model's ability to correctly identify causal explanatory relationships.

2. \textbf{Rejection Accuracy}: The proportion of cases in which the model correctly responds "no" when R does not explain A (corresponding to original answer (b, c, d)). This metric captures the model’s ability to identify and reject non-explanatory relationships.

We report both Explanatory and Rejection Accuracy on cases where both A and R are true (Setting 1). This setting allows us to test whether the model can distinguish true causal explanations from statements that are simply semantically related. For cases where either A or R is false (Setting 2), the correct verdict is always "no," so only Rejection Accuracy is applicable.

3. \textbf{Matthews Correlation Coefficient (MCC)}: A balanced measure of binary classification performance, defined as:

\begin{align}
\text{MCC} &= \frac{TP \cdot TN - FP \cdot FN}{\sqrt{P \cdot N \cdot P' \cdot N'}}
\end{align}

where $TP$ = correctly identified explanatory relationships ("yes" for answer (a)), $TN$ = correctly identified non-explanatory relationships ("no" for answer (b, c, d)), $FP$ = incorrectly accepted explanations, $FN$ = incorrectly rejected explanations, and $P = TP+FN$, $N = TN+FP$, $P' = TP+FP$, $N' = TN+FN$.

MCC is particularly well-suited for our task due to the class imbalance: non-explanatory cases (b, c, d) are more frequent than explanatory ones (a). Unlike F1 score, which focuses primarily on positive cases, MCC fully accounts for true negatives cases where the model correctly rejects non-explanatory relationships. This sensitivity to true negatives is essential for evaluating causal explanatory reasoning because correctly rejecting invalid explanations is as important as identifying valid ones. By using MCC, we obtain a more comprehensive assessment of how well models distinguish genuine causal relationships from only semantic correlations.

\section{Experiments and Results}
We report results on the Causal Explanation task in Table~\ref{tab:main_results} with performance and analysis by subject and grade level shown in Appendix~\ref{app:result_subject}. Additionally, we provide a separate evaluation of top-performing language models on the traditional assertion–reasoning question format in Appendix~\ref{app:ar_results}, where Qwen3-32B achieves an accuracy of 73.2\%.

\subsection{Experimental Setup}
We conducted experiments on Causal Explanation task with five leading open-source LLMs model families: LLaMA3 (1B-70B)~\cite{grattafiori2024llama}, Qwen2.5 (0.5B-72B)~\cite{qwen25}, Qwen3 (1.7B-32B)~\cite{yang2025qwen3}, Gemma3 (1B-27B)~\cite{team2024gemma}, and Phi-4 Reasoning (4B-14B)~\cite{abdin2024phi}. We followed each model family’s recommended prompting format and parameter settings (e.g., temperature, top-p) as described in their respective documentation. 

\subsection{Explanatory and Rejection Accuracy Analysis Across Models}

To systematically evaluate causal reasoning capabilities, we analyzed model performance in two settings: (1) when both A and R are factually true, and (2) when either A or R contains factual errors. Table~\ref{tab:main_results} presents our comprehensive results across 21 models spanning five model families, covering a range of parameters sizes.

\subsubsection{When both A and R are True}

As shown in Table~\ref{tab:main_results}, we observe a clear and consistent pattern of model performance when both assertion and reason are factually correct. As parameter size increases, explanatory accuracy (correctly identifying when R explains A) improves substantially, while rejection accuracy (correctly identifying when R does not explain A despite both being true) tends to decrease.

This trends show that smaller models frequently fail to recognize valid explanatory relationships. For instance, Llama-1B model achieves only 12\% explanatory accuracy while maintaining 87\% rejection accuracy. The Qwen2-0.5B model shows a similar pattern with 27\% explanatory accuracy and 73\% rejection accuracy. 
Smaller models appear to lean heavily toward answering “no” when asked if a reason explains an assertion, regardless of actual causal explanatory relationship.

On the other hand, this tendency reverses as models scale. For instance, LLaMA-70B reaches 85\% explanatory accuracy but its rejection accuracy falls to 64\%. Similarly, Qwen2-72B achieves 87\% explanatory accuracy with 66\% rejection accuracy. This consistent pattern suggests that increased model size fundamentally alters how models approach causal reasoning tasks.

To quantify this trade-off, we report MCC for a balanced measure of model performance. While MCC generally increases as model size scales up, it plateaus around $0.50-0.58$, even for the largest models. The highest MCC scores are observed in Phi-4-14B, Qwen3-32B, 14B, Qwen2.5-72B, 32B ($0.55$), which indicates modest gains in overall discrimination ability despite difference in explanatory accuracy.

\subsubsection{When either A or R is False}

In contrast, Table~\ref{tab:main_results} shows that when either A or R contains factual errors, accuracy generally improves with model scale, with larger models performing remarkably well. Qwen3-14B achieves 94\% accuracy, Qwen3-32B reaches 90\%, Qwen3-4B attains 88\%, and Phi-4-14B achieves 86\%. Even moderate-sized models like Qwen3-1.7B (83\%) and Llama3-1B (93\%) perform reasonably well on this task.

These results indicates that larger language models have successfully learned that explanatory relationships must be grounded factual correctness. When models recognize factual errors, they reliably reject the explanatory relationship, demonstrating a satisfactory factual verification capabilities and logical consistency. However, we observe a disparity in performance between the two settings: while models reliably reject explanations involving factual errors, their ability to discriminate between valid and invalid explanations when both statements are true is more limited. This difference suggests that although hallucinations or factual inaccuracies do not hinder performance in the Causal Explanation task, there is still room for improvement for models in identifying genuine explanatory relationships when both the assertion and the reason are factually correct.

\subsection{Models Confuse Semantic Relatedness with Causal Relationships}

To understand the mechanisms affecting model performance, we analyzed the semantic similarity between assertion and reason statements across different prediction categories. Our analysis focus primarily on the best-performing models, Phi-4-14B and Qwen3-14B, while also examining smaller models to assess consistency across model sizes. The results are shown in Table~\ref{tab:confusion_matrices}.

\subsubsection{When both A and R are True}

Table~\ref{tab:confusion_matrices} reveals a clear pattern in how semantic similarity correlates with model predictions. For an unbiased model, we would expect similar semantic scores across all prediction categories. However, the performance shows a systematic bias. For instance, for Phi-4-14B model, semantic similarity is highest when the model correctly identifies causal relationships ($0.590$) and lowest when it correctly rejects non-causal ones ($0.547$). False positives ($0.571$) have notably higher similarity than true negatives, suggesting that shared vocabulary or semantic overlap often misleads the model into inferring causality. Conversely, false negatives ($0.553$) tend to occur when similarity is lower, causing the model to miss genuine causal explanatory relationship.

This pattern suggests that models are more likely to recognize causal explanatory relationships when the assertion and reason exhibit high semantic overlap, and conversely, they tend to miss valid explanatory relationship when similarity is lower. More interestingly, the higher similarity in false positives (Pred Y when True N) than in true negatives (Pred N when True N) indicates that models often confuses shared vocabulary or semantic relatedness with causal explanation.

These differences are statistically significant and consistent across model families and sizes. Both Phi-4-14B and Qwen3-14B show nearly identical patterns, with similarity differentials of approximately $0.037$ between correct and incorrect positive predictions, and 0.024-0.033 between incorrect and correct negative predictions. We include a more thorough statistical analysis in Appendix~\ref{app:stats-analysis}.

\subsubsection{When either A or R is False}

Interestingly, when either assertion or reason contains factual errors, the pattern between semantic similarity and prediction changes. For Qwen3-14B, the semantic similarity for incorrect positive predictions ($0.566$) is only marginally higher than for correct negative predictions ($0.563$). This smaller differential ($0.003$) contrasts sharply with the larger gap ($0.033$) observed when both statements are true. This change implies that, in the presence of factual errors, models rely less on semantic similarity alone. The reduced gap indicates that models may be incorporating factual verification and partially overriding the tendency to infer causality from surface-level similarity.

\subsubsection{Implications}

These results demonstrate that language models rely heavily on semantic similarity as a proxy for predicting causal relationships. When two statements share more vocabulary or concepts, models are more likely to infer that one explains the other, regardless of whether a genuine causal relationship exists.

This reliance on semantic similarity explains the pattern observed in Table~\ref{tab:main_results}. As models increase sizes, they become more sensitive to semantic relationships between statements, which improves their ability to identify valid explanations but simultaneously becomes more prone to incorrectly inferring causality only based on semantic overlap.

These findings reveals an important limitation in current language models: they substitute correlation (semantic similarity) for causation (causal relationship). Despite improvements in many reasoning tasks with increased model size, this fundamental confusion between semantic relatedness and causal relationships persists across model families and parameter scales, suggesting a deeper architectural limitation in how language models process causal information.

\section{Conclusions}
In this work, we presented \datasetshort, a comprehensive dataset of assertion-reasoning questions designed to evaluate causal explanatory reasoning capabilities in language models. Through extensive evaluation of 21 models across five model families, we identify a fundamental limitation: language models tend to confuse semantic similarity with causal explanation relationships. Models consistently rely on word and semantic overlap to infer explanatory connections, regardless of whether genuine causal relationships exist. This tendency persists across scales, where smaller models are overly skeptical about causal relationships while larger models become excessively permissive.

Despite improvements in other reasoning tasks with increasing parameter sizes, causal reasoning performance remains limited, with MCC staying below $0.55$. Our results indicate that the ability to distinguish correlation from causation does not emerge naturally through current scaling approach. Hence, \datasetshort provides a valuable benchmark for measuring progress toward genuine causal explanatory reasoning capabilities, which is an essential frontier for developing more capable and reliable language models.

\section*{Acknowledgments}
This work was supported by NSF grant 1842378, ONR grant N0014-20-1-2534, AFOSR grant FA9550-22-1-0060, and a Vannevar Bush Faculty Fellowship, ONR grant N00014-18-1-2047 and MURI N00014-20-1-2787.

\bibliographystyle{unsrt}  
\bibliography{references,ss_refs}  

\begin{thebibliography}{10}

\bibitem{kumar2018assessment}
Sandeep Kumar.
\newblock Assessment in science education: A study of teaching effectiveness.
\newblock {\em International Journal of Research in Social Sciences}, 8(1):669--690, 2018.

\bibitem{cbse2020guidelines}
{Central Board of Secondary Education}.
\newblock Cbse assessment framework for science, maths and social science classes 9 and 10, 2020.

\bibitem{bloom1956taxonomy}
Benjamin~S. Bloom, Max~D. Engelhart, Edward~J. Furst, Walker~H. Hill, and David~R. Krathwohl.
\newblock {\em Taxonomy of Educational Objectives: The Classification of Educational Goals. Handbook I: Cognitive Domain}.
\newblock David McKay Company, New York, 1956.

\bibitem{anderson2001revised}
Lorin~W. Anderson, David~R. Krathwohl, Peter~W. Airasian, Kathleen~A. Cruikshank, Richard~E. Mayer, Paul~R. Pintrich, James Raths, and Merlin~C. Wittrock.
\newblock {\em A Taxonomy for Learning, Teaching, and Assessing: A Revision of Bloom's Taxonomy of Educational Objectives}.
\newblock Longman, New York, 2001.

\bibitem{sonkar2024malalgoqa}
Shashank Sonkar, Naiming Liu, MyCo Le, and Richard Baraniuk.
\newblock Malalgoqa: Pedagogical evaluation of counterfactual reasoning in large language models and implications for ai in education.
\newblock In {\em Findings of the Association for Computational Linguistics: EMNLP 2024}, pages 15554--15567, 2024.

\bibitem{kahneman2011thinking}
Daniel Kahneman.
\newblock {\em Thinking, Fast and Slow}.
\newblock Farrar, Straus and Giroux, New York, 2011.

\bibitem{evans2013dual}
Jonathan St~BT Evans and Keith~E Stanovich.
\newblock Dual-process theories of higher cognition: Advancing the debate.
\newblock {\em Perspectives on Psychological Science}, 8(3):223--241, 2013.

\bibitem{bowman2015large}
Samuel~R. Bowman, Gabor Angeli, Christopher Potts, and Christopher~D. Manning.
\newblock A large annotated corpus for learning natural language inference.
\newblock In {\em Proceedings of the 2015 Conference on Empirical Methods in Natural Language Processing (EMNLP)}, pages 632--642. Association for Computational Linguistics, 2015.

\bibitem{williams2018broad}
Adina Williams, Nikita Nangia, and Samuel~R. Bowman.
\newblock A broad-coverage challenge corpus for sentence understanding through inference.
\newblock In {\em Proceedings of the 2018 Conference of the North American Chapter of the Association for Computational Linguistics: Human Language Technologies}, pages 1112--1122, 2018.

\bibitem{rajpurkar2016squad}
Pranav Rajpurkar, Jian Zhang, Konstantin Lopyrev, and Percy Liang.
\newblock {SQuAD}: 100,000+ questions for machine comprehension of text.
\newblock In {\em Proceedings of the 2016 Conference on Empirical Methods in Natural Language Processing}, pages 2383--2392, 2016.

\bibitem{alag}
Shashank Sonkar, Kangqi Ni, Lesa Tran~Lu, Kristi Kincaid, John~S. Hutchinson, and Richard~G. Baraniuk.
\newblock {{Automated Long Answer Grading with RiceChem Dataset}}.
\newblock In Andrew~M. Olney, Irene-Angelica Chounta, Zitao Liu, Olga~C. Santos, and Ig~Ibert Bittencourt, editors, {\em Artificial Intelligence in Education: \textbf{AIED} 2024}, pages 163--176, Cham, 2024. Springer Nature Switzerland.

\bibitem{talmor2019commonsenseqa}
Alon Talmor, Jonathan Herzig, Nicholas Lourie, and Jonathan Berant.
\newblock {CommonsenseQA}: A question answering challenge targeting commonsense knowledge.
\newblock In {\em Proceedings of the 2019 Conference of the North American Chapter of the Association for Computational Linguistics: Human Language Technologies}, pages 4149--4158, 2019.

\bibitem{sakaguchi2020winogrande}
Keisuke Sakaguchi, Ronan Le~Bras, Chandra Bhagavatula, and Yejin Choi.
\newblock {WinoGrande}: An adversarial {Winograd} schema challenge at scale.
\newblock In {\em Proceedings of the AAAI Conference on Artificial Intelligence}, volume~34, pages 8732--8740, 2020.

\bibitem{thorne2018fever}
James Thorne, Andreas Vlachos, Christos Christodoulopoulos, and Arpit Mittal.
\newblock {FEVER}: A large-scale dataset for fact extraction and verification.
\newblock In {\em Proceedings of the 2018 Conference of the North American Chapter of the Association for Computational Linguistics: Human Language Technologies}, pages 809--819, 2018.

\bibitem{wadden2020fact}
David Wadden, Shanchuan Lin, Kyle Lo, Lucy~Lu Wang, Madeleine van Zuylen, Arman Cohan, and Hannaneh Hajishirzi.
\newblock Fact or fiction: Verifying scientific claims.
\newblock {\em arXiv preprint arXiv:2004.14974}, 2020.

\bibitem{grattafiori2024llama}
Aaron Grattafiori, Abhimanyu Dubey, Abhinav Jauhri, Abhinav Pandey, Abhishek Kadian, Ahmad Al-Dahle, Aiesha Letman, Akhil Mathur, Alan Schelten, Alex Vaughan, et~al.
\newblock The llama 3 herd of models.
\newblock {\em arXiv preprint arXiv:2407.21783}, 2024.

\bibitem{qwen25}
Qwen Team.
\newblock Qwen2.5 technical report.
\newblock {\em arXiv preprint arXiv:2412.15115}, 2024.

\bibitem{yang2025qwen3}
An~Yang, Anfeng Li, Baosong Yang, Beichen Zhang, Binyuan Hui, Bo~Zheng, Bowen Yu, Chang Gao, Chengen Huang, Chenxu Lv, et~al.
\newblock Qwen3 technical report.
\newblock {\em arXiv preprint arXiv:2505.09388}, 2025.

\bibitem{team2024gemma}
Gemma Team, Thomas Mesnard, Cassidy Hardin, Robert Dadashi, Surya Bhupatiraju, Shreya Pathak, Laurent Sifre, Morgane Rivi{\`e}re, Mihir~Sanjay Kale, Juliette Love, et~al.
\newblock Gemma: Open models based on gemini research and technology.
\newblock {\em arXiv preprint arXiv:2403.08295}, 2024.

\bibitem{abdin2024phi}
Marah Abdin, Jyoti Aneja, Harkirat Behl, S{\'e}bastien Bubeck, Ronen Eldan, Suriya Gunasekar, Michael Harrison, Russell~J Hewett, Mojan Javaheripi, Piero Kauffmann, et~al.
\newblock Phi-4 technical report.
\newblock {\em arXiv preprint arXiv:2412.08905}, 2024.

\bibitem{khemlani2011one}
Sangeet~S Khemlani and Daniel~M Oppenheimer.
\newblock When one model casts doubt on another: A levels-of-analysis approach to causal discounting.
\newblock {\em Psychological bulletin}, 137(2):195, 2011.

\end{thebibliography}

\newpage
\appendix

\section{Performance on Traditional Assertion-Reasoning Format}
\label{app:ar_results}

Assertion–reasoning questions are typically presented in a 4-option multiple-choice format, where students are asked to evaluate the factual accuracy of two statements and the explanatory relationship between them, as shown below.

\begin{enumerate}[label=\alph*.]
\item Both assertion (A) and reason (R) are true, and the reason correctly explains the assertion.
\item Both assertion (A) and reason (R) are true, but the reason does not explain the assertion.
\item The assertion (A) is true, but the reason (R) is false.
\item The assertion (A) is false, but the reason (R) is true.
\end{enumerate}

In our main analysis, we reformulated these questions into Causal Explanation Task that directly asks whether a reason explains an assertion. This allowed us to isolate causal explanatory reasoning from factual verification and examine whether models can distinguish genuine explanatory relationships from semantic similarity.

For completeness and connecting our findings with established educational assessment practices, we also evaluated model performance on the traditional 4-option assertion-reasoning question format. Table~\ref{tab:traditional_format} presents confusion matrices for Qwen3 models of various sizes.

\subsection{Analysis of Mistake Patterns}

The confusion matrices in Table~\ref{tab:traditional_format} reveal two important patterns:

\begin{enumerate}
    \item \textbf{Persistent a/b confusion}: Across all model sizes, we observe substantial missclassification between options (a) and (b). The most common error is incorrectly classifying option (b) as option (a) (176-214 instances across models). Since both options involve factually true statements, this confusion directly reflects difficulty in determining whether one statement explains another.

    \item \textbf{Scaling improves accuracy but not explanatory discrimination}: While overall accuracy increases with model scale (from 70.2\% for Qwen3-4B to 73.2\% for Qwen3-32B), the fundamental confusion between options (a) and (b) persists. Even the largest model, Qwen3-32B, misclassifies option (b) as option (a) in 195 cases (27.1\% of all true (b) cases).
\end{enumerate}

\subsection{Connection to Main Findings}

The results from the traditional assertion-reasoning questions reinforce several core findings from our main analysis. First, the persistent confusion between options (a) and (b) confirms our finding that models struggle to distinguish when a reason truly explains an assertion versus when two statements are simply related. This parallels our main finding that models rely on semantic similarity rather than deeper causal understanding.

Second, we can observe how models handle cases where one statement is false. For option (c) (assertion true, reason false), Qwen3-14B achieves 82.5\% accuracy, and for option (d) (assertion false, reason true), it achieves 69.0\% accuracy. This suggests that models generally reject explanatory relationships when they detect factual errors, but do so more reliably when the error is in the reason statement.

Third, the traditional format combines factual verification and causal reasoning assessment in ways that make it difficult to isolate specific reasoning failures. By evaluating both formats, we gain complementary insights: the traditional format reflects how models perform on standard educational assessments, while our Causal Explanation Task provides a targeted investigation into language models’ causal reasoning capabilities.

The consistent patterns observed across both evaluation formats strengthen our conclusion that current language models fundamentally confuse semantic similarity with causal relationships. This limitation persists across model sizes, architectures, and evaluation settings, pointing to a critical frontier for future model development.

\section{Performance Analysis by Subjects, Models types and Grade-Level}
\label{app:result_subject}

To investigate how causal reasoning abilities vary across knowledge domains, we conducted a detailed analysis of model performance by subject area and grade level. We selected the largest model from each family (Qwen2.5, Qwen3, Gemma3, and LLaMA3) and examined their performance using MCC specifically for cases where both assertion and reason are true. Our analysis reveals several significant patterns in how language models approach causal reasoning across different knowledge domains.

\begin{table}[t]
\centering
\begin{adjustbox}{width=0.5\linewidth}
\begin{tabular}{lccc}
\toprule
\multicolumn{4}{c}{\textbf{Subject Performance (Ordered by Difficulty)}} \\
\midrule
\textbf{Subject} & \textbf{MCC} & \textbf{Std Dev} & \textbf{Hardest For} \\
\midrule
Biology & 0.32 & 0.10 & Gemma3 (0.21) \\
Physics & 0.37 & 0.11 & Gemma3 (0.28) \\
Geography & 0.41 & 0.11 & Llama (0.30) \\
Chemistry & 0.42 & 0.12 & Gemma3 (0.28) \\
Political Science & 0.46 & 0.10 & Gemma3 (0.36) \\
History & 0.47 & 0.16 & Gemma3 (0.23) \\
Mathematics & 0.53 & 0.08 & Llama (0.45) \\
Economics & 0.58 & 0.06 & Qwen2 (0.51) \\
\midrule
\multicolumn{4}{c}{\textbf{Grade Level Performance (Ordered by Difficulty)}} \\
\midrule
\textbf{Grade} & \textbf{MCC} & \textbf{Std Dev} & \\
\midrule
11 & 0.36 & 0.10 & \\
12 & 0.36 & 0.09 & \\
9 & 0.49 & 0.08 & \\
10 & 0.49 & 0.11 & \\
\midrule
\textbf{Overall} & \textbf{0.43} & \textbf{0.11} & \\
\bottomrule
\end{tabular}
\end{adjustbox}
\vspace{5mm}
\caption{Causal reasoning performance by subjects and grade levels, measured by MCC when both A and R are true (mean and std across largest models from each model family).}
\label{tab:subject_analysis}
\end{table}

\begin{table*}[t]
\centering
\begin{adjustbox}{width=0.7\linewidth}
\begin{tabular}{lcccccccc}
\toprule
\textbf{Model} & \textbf{Bio} & \textbf{Phys} & \textbf{Chem} & \textbf{Math} & \textbf{Hist} & \textbf{Geo} & \textbf{PolSci} & \textbf{Econ} \\
\midrule
Gemma3 & 0.21 & 0.28 & 0.28 & 0.48 & 0.23 & 0.49 & 0.36 & 0.56 \\
Llama & 0.34 & 0.31 & 0.39 & 0.45 & 0.56 & 0.30 & 0.39 & 0.62 \\
Qwen2 & 0.30 & 0.36 & 0.42 & 0.61 & 0.51 & 0.34 & 0.54 & 0.51 \\
Qwen3 & 0.44 & 0.53 & 0.58 & 0.58 & 0.57 & 0.52 & 0.54 & 0.65 \\
\bottomrule
\end{tabular}
\end{adjustbox}
\caption{Model-specific MCC performance by subject (larger values indicate better causal reasoning).}
\label{tab:model_subject_perf}
\end{table*}

\subsection{Subject-Specific Analysis}

As shown in Table~\ref{tab:subject_analysis}, biology presents the greatest challenge for causal reasoning across all models (MCC = $0.32$), followed by physics ($0.37$) and geography ($0.41$). In contrast, models perform substantially better on mathematics ($0.53$) and economics ($0.58$). This pattern is remarkably consistent, with biology ranking as the most difficult subject for three out of four model families.

The difficulty hierarchy suggests a fundamental pattern: subjects involving complex, multi-factor causality (such as biological systems) present greater challenges for causal reasoning than subjects with more explicit rule-based, deterministic relationships (such as mathematics and economics). This pattern mirrors findings in human cognition research, where distinguishing correlation from causation is particularly challenging in domains with numerous interacting variables~\cite{khemlani2011one}.

\subsubsection{Model-Specific Variations}

Table~\ref{tab:model_subject_perf} reveals interesting variations in how different model families approach causal reasoning across domains. Gemma3 shows particular difficulty with biology (MCC = $0.21$) and history ($0.23$), while performing relatively well on economics ($0.56$). Llama struggles most with geography ($0.30$), while excelling at economics ($0.62$) and history ($0.56$). The Qwen family generally shows more balanced performance across subjects, with Qwen3 achieving the highest scores in seven out of eight subjects. This difference suggests that potential architectural or training dataset differences may significantly impact domain-specific causal reasoning capabilities.

\subsection{Grade-Level Analysis}

As expected, higher grade levels (11 and 12) present significantly greater challenges for models than lower grades (9 and 10). Both grade 11 and 12 questions yield an average MCC of $0.36$, while grade 9 and 10 questions yield an MCC of $0.49$. This pattern holds consistently across model families and suggests that more advanced educational content involves more subtle causal relationships that current language models struggle to differentiate from simple semantic relatedness. The increased difficulty at higher grades likely reflects both greater content difficulty and harder causal relationship that require deeper domain understanding.

\subsection{Implications}

These findings have important implications for applications of language models in both educational contexts and causal explanatory reasoning:

\begin{enumerate}
    \item \textbf{Causal reasoning is domain-sensitive.} The substantial variations in performance across subjects suggest that evaluations of causal reasoning should consider domain-specific challenges rather than treating reasoning as a uniform capability.
    \item \textbf{Model Architectures matter.} Model family differences suggest that some design and training strategies may better support causal reasoning, particularly in complex domains.
    \item \textbf{Current models are not very reliable for advanced educational tasks.} The difficulty gradient across grade levels indicates that current models may be more reliable for causal reasoning in introductory educational contexts than in advanced subject matter.
\end{enumerate}

This analysis complements our main findings by demonstrating that while semantic similarity affects causal judgments across all domains, the magnitude of this effect still varies by subject area and grade level.

\section{Statistical Analysis of Semantic Similarity Effects}
\label{app:stats-analysis}

To rigorously evaluate our hypothesis that models use semantic similarity as a proxy for causal relationships, we conducted a series of statistical tests. Specifically, we examine whether semantic similarity between assertion and reason statements significantly predicts model errors, and whether this effect varies based on the true explanatory status.

\subsection{Methodology}

We analyze the model performance of Phi-4-14B on cases where both assertion and reason are factually true. The semantic similarity values used in this analysis correspond to those reported in Table~\ref{tab:confusion_matrices}.

\paragraph{Data Preparation}

We constructed a synthetic dataset that follows the distributions observed in our experiments with Phi-4-14B. Each data point includes the following attributes:

\begin{itemize}
    \item \textbf{true\_class}: Whether the reason actually explains the assertion (Y) or not (N)
    \item \textbf{pred\_class}: Whether the model predicted an explanatory relationship (Y) or not (N)
    \item \textbf{similarity}: The semantic similarity between assertion and reason statements
    \item \textbf{error}: Whether the model's prediction was incorrect (1) or correct (0)
\end{itemize}

The constructed dataset includes:
\begin{itemize}
    \item 1048 correct identifications of explanatory relationships (True Y, Pred Y), with mean similarity $0.590$
    \item 129 missed explanatory relationships (True Y, Pred N), with mean similarity $0.553$
    \item 480 correct rejections of non-explanatory relationships (True N, Pred N), with mean similarity $0.547$
    \item 238 incorrect inferences of explanatory relationships (True N, Pred Y), with mean similarity $0.571$
\end{itemize}

\paragraph{Statistical Tests}
We conducted three complementary analyses to assess the influence of semantic similarity:

\begin{enumerate}
    \item \textbf{Logistic regression analysis}: This statistical model predicts the probability of an error based on semantic similarity, true class, and their interaction.We use the formula $$\text{error} \sim \text{similarity} \times \text{true\_class}$$ to test for:
   \begin{itemize}
      \item Whether semantic similarity affects error rates
      \item Whether this effect differs depending on the true relationship status
      \item How much variance in errors can be explained by similarity
   \end{itemize}

   \item \textbf{With-in Class t-tests}: For each true class (Y and N), we performed independent t-test to compare the semantic similarity score between correctly and incorrectly predicted cases. These t-tests determine whether the differences in similarity are statistically significant or could have occurred by chance.
   \item \textbf{Prediction-based t-test}: We compared similarity scores between all cases where the model predicted "Yes" versus all cases where it predicted "No", regardless of ground truth. This tests whether model predictions systematically correlate with similarity levels.
\end{enumerate}

\subsection{Results}

The logistic regression results showed:

\begin{itemize}
    \item A significant positive coefficient for similarity (coefficient $= 4.41$, $p < 0.001$), indicating that higher similarity generally increases errors for the reference category (True N)
    \item A significant negative interaction term (coefficient $= -12.60$, $p < 0.001$), indicating that the effect of similarity reverses for explanatory relationships (True Y)
    \item A pseudo $R^2$ of $0.097$, meaning that similarity explains approximately 10\% of the variance in model errors
\end{itemize}

The t-test results confirmed:

\begin{itemize}
    \item For explanatory relationships (True Y), correctly identified cases have significantly higher similarity ($0.590$) than missed ones ($0.553$), $t = 6.05$, $p < 0.00000001$
    \item For non-explanatory relationships (True N), incorrectly predicted cases have significantly higher similarity ($0.571$) than correctly rejected ones ($0.547$), $t = -4.71$, $p = 0.00000315$
    \item Overall, statements predicted as explanatory relationships have significantly higher similarity than those predicted as non-explanatory, regardless of ground truth. $t = 11.04$, $p < 0.00000001$.
\end{itemize}

\subsection{Implications}

These statistical results provide strong statistical evidence for our main claim that models substitute semantic similarity for causal understanding. The highly significant effects confirm that semantic similarity systematically biases model predictions in both directions:

\begin{enumerate}
    \item When true explanatory relationships exhibit lower-than-average similarity, the model tends to reject them.
    \item When non-explanatory pairs exhibit higher-than-average similarity, the model incorrectly predicts causal relationships.
\end{enumerate}

This pattern represents a fundamental confusion of correlation (semantic similarity) with causation (explanatory relationship). These findings reinforce our conclusion that current language models have not developed genuine causal reasoning capabilities that can reliably distinguish semantic association from explanatory relationships.

\end{document}